# AN IMPLEMENTATION, EMPIRICAL EVALUATION AND PROPOSED IMPROVEMENT FOR BIDIRECTIONAL SPLITTING METHOD FOR ARGUMENTATION FRAMEWORKS UNDER STABLE SEMANTICS

Renata Wong[1,2]

[1]Department of Computer Science and Technology, Nanjing University, Nanjing, China
[2]State Key Laboratory for Novel Software Technology at Nanjing University, Nanjing, China.

## ABSTRACT

*Abstract argumentation frameworks are formal systems that facilitate obtaining conclusions from non-monotonic knowledge systems. Within such a system, an argumentation semantics is defined as a set of arguments with some desired qualities, for example, that the elements are not in conflict with each other. Splitting an argumentation framework can efficiently speed up the computation of argumentation semantics. With respect to stable semantics, two methods have been proposed to split an argumentation framework either in a unidirectional or bidirectional fashion. The advantage of bidirectional splitting is that it is not structure-dependent and, unlike unidirectional splitting, it can be used for frameworks consisting of a single strongly connected component. Bidirectional splitting makes use of a minimum cut. In this paper, we implement and test the performance of the bidirectional splitting method, along with two types of graph cut algorithms. Experimental data suggest that using a minimum cut will not improve the performance of computing stable semantics in most cases. Hence, instead of a minimum cut, we propose to use a balanced cut, where the framework is split into two sub-frameworks of equal size. Experimental results conducted on bidirectional splitting using the balanced cut show a significant improvement in the performance of computing semantics.*

## 1. INTRODUCTION

Nowadays, much research in the field of argumentation relates to the notion of an *abstract argumentation framework (AF)* introduced by Phan Minh Dung in 1995 [1-4]. Essentially, an AF can be conceptualized as a directed graph with arguments represented by vertices, and conflicts between arguments represented by directed edges. Being dissociated from any specific instance, an abstract framework is thus able to cover a variety of situations and can be instantiated for use in various empirical fields. It also allows for various extensions to improve its expressiveness and performance. Some of the proposed extensions include e.g. value-based argumentation frameworks by Bench-Capon et al [5], logic-based argumentation frameworks by Besnard and Hunter [6], probabilistic argumentation frameworks by Li et al [7-8], or preference-based argumentation frameworks by Amgoud and Cayrol [9].

In 1994, Lifschitz and Turner [10] presented the notion of splitting a logic program. In general terms, splitting involves dividing a program into two parts and computing each part separately with the hope that it will lead to a computational speedup. As suggested by the authors, splitting worked especially well for programs with negation as failure. In his seminal paper of 1995, Dung [1] suggested that argumentation can be viewed as a form of logic programming with negation as failure. The research by Caminada et al [11] indicates the existence of certain equivalences between the semantics for logic programming and the semantics for abstract argumentation. Based on this analogy, a splitting procedure for Dung-style argumentation frameworks was developed in [12] (hereafter referred to as *unidirectional splitting*) and subsequently empirically evaluated in [13].

In unidirectional splitting, the framework is separated along edges that all point into the same direction. This method is applicable to frameworks consisting of at least two strongly connected components (SCC). In order to overcome this limitation, a method of *bidirectional splitting* was proposed in [14] where it is called *parameterized splitting*. In bidirectional splitting, the framework is separated along edges that do not have to point into the same direction. Bidirectional splitting also works for frameworks with an arbitrary number of SCCs.

Standard argumentation semantics proposed by P. M. Dung include complete, preferred, grounded and stable semantics. In this study we implement and experimentally evaluate the bidirectional splitting method given in the theory paper [13] which was developed specifically for stable semantics. The desirability of speeding up the computation of stable semantics stems from the well known fact that the question of whether a stable argument set exists for a given argumentation framework is an NP-complete problem (see e.g. [15]).

Our experimental results demonstrate that splitting leads to improvement in computational performance for stable semantics. The speedup is slight if a minimum cut algorithm is applied together with the splitting procedure, and tends to decrease with the complexity of the framework. We show that dividing a framework into halves (hereafter referred to as *balanced cut*, or BC) has a better run time than applying a minimum cut.

The algorithm for bidirectional splitting is a constituent part of an argumentation framework editor, a standalone application we developed in the Java programming language. Experimental evaluation is conducted using this application.

The rest of this paper is structured as follows: Sections 2 and 3 present the theoretical background of argumentation frameworks and of splitting, respectively. Section 4 describes the setup of the conducted experiments, while the experimental results and related discussion are given in Section 5. The concluding remarks are presented in Section 6, together with a summary of the proposed improvement. The paper ends with the Acknowledgments.

## 2. ARGUMENTATION FRAMEWORKS AND SEMANTICS

An argumentation framework $F = (A, R)$ is a pair in which $A$ is a non-empty finite set of elements called *arguments*, and $R \subseteq (A \times A)$ is a binary relation defined on $A$ called *attack relation*.

Certain subsets of the argument set called *admissible* sets are of special interest. An admissible subset $S \subseteq A$ is a set with the following characteristics:

(1) $S$ is free of conflicts, i.e. $\sim \exists\, a, b \in S$ such that $(a, b) \in R$, and

(2) every argument $a \in S$ is *acceptable* with regard to $S$, i.e. $\forall b \in A: (b, a) \in R \rightarrow \exists c \in S \land (c, b) \in R$.

Argumentation semantics is an admissible set with additional features. In the case of stable semantics it is required that every argument acceptable with regard to $S$ is also included in $S$ and that any argument not in $S$ is attacked by some argument in $S$. In essence, this is the definition of a stable extension. Another approach to semantics is labeling where we have three subsets called *in, out, undec* (here we follow the formalism given in [16]) instead of specifying only the subset $S$ that has the required features. An argument is labeled *in* if it is not attacked or if it is attacked by an argument labeled *out*. An argument is labeled *out* if it is attacked by at least one argument with the label *in*. Any other argument is labeled *undec*. An extension corresponds to the set *in*.

In order to introduce the concept of a stable labeling, we need to define the notion of legality of labels:

**Definition 1** ([17])

Let L be a labeling for an argumentation framework $F = (A, R)$. We call $a \in A$:
  (1) *legally in* iff $a \in in(L)$ and $\forall b: (b, a) \in R \rightarrow b \in out(L)$
  (2) *legally out* iff $a \in out(L)$ and $\exists b: (b, a) \in R \land b \in in(L)$
  (3) *legally undec* iff (1) $a \in undec(L)$, (2) $\sim \forall b: (b, a) \in R \rightarrow b \in out(L)$, and (3) $\sim \exists b: (b, a) \in R \land b \in in(L)$
  (4) *illegally lab* for a given $lab \in \{in, out, undec\}$ iff $a \in lab(L)$ and $a$ is not *legally lab*

A stable labeling is usually defined in the literature on the basis of a complete labeling. For simplicity, this paper incorporates the notion of complete labeling into that of stable labeling.

**Definition 2** (Stable Labeling [12])

A *stable labeling* $L_S$ of an argumentation framework $F = (A, R)$ is a labeling that does not contain any arguments that are *illegally in, illegally out*, or *legally undec*, and where $undec(L) = \emptyset$.

Stable semantics finds its exact correspondence in such non-monotonic formalisms as Moore's autoepistemic logic, Reiter's default logic, or logic programming [18, 1]. It is to be noted that not every AF possesses a stable extension or labeling. This issue was addressed by Caminada and Dunne [19] by proposing the notion of a *semi-stable semantics*. The classical stable semantics and the semi-stable semantics are related in such a way that every stable labeling is also a semi-stable labeling. Thus, if a stable labeling exists, it will correspond to a semi-stable labeling. Furthermore, the existence of at least one semi-stable set is guaranteed for any finite argumentation framework.

**Example 1**: The AF shown in Figure 1 has one stable extension/labeling. As argument 4 has no attackers, it is labeled *in* and therefore constitutes an element of the stable extension. Being attacked by an argument labeled *in*, arguments 0 and 3 are thus both labeled *out*. Having an attacker that is labeled *out*, argument 1 is labeled *in* and belongs thus to the extension. The last argument, 2, is also labeled *out* as it is attacked by an argument labeled *in*. Hence, the stable extension of this example framework is {1,4}, and the stable labeling is {{1,4},{0,2,3},∅}.

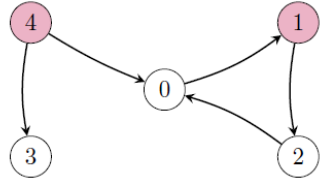

Figure 1: Example of an argumentation framework with its stable labelling

As mentioned in our introduction, the question whether a stable extension exists in a given AF belongs to the class of NP-complete problems. It is thus desirable to devise methods that have the potential to improve the computational performance for stable semantics. There are several such possible approaches. (1) Greco and Parisi [20] propose a method for recomputing grounded semantics and ideal semantics [21] in the case that the AF was modified. It involves a recalculation of only those arguments that have been affected by the modification. To our knowledge, no such method has yet been proposed for stable semantics. (2) A parallel algorithm has been proposed by Finkel et al [22] for computing stable models in logic programs. Again, to our knowledge, no parallel method for stable semantics in argumentation has yet been devised, although the correspondence between stable models in logic programming and stable semantics in argumentation have been identified in [1]. (3) Doumbouya et al [23] approach the problem by analyzing the structure of the underlying graph. Their study proves that if the graph is asymmetric and irreflexive and if there exists an argument that attacks all the others then there is a unique stable extension formed by that argument. The method is helpful for predicting if a stable extension exists, further research on the graph structure however is needed to specify the exact conditions that govern the existence of a stable extension/labeling in a given AF. (4) Yet another attempt is [13] where the problem of computing stable semantics in argumentation is addressed by means of splitting a framework in order to compute the semantics more efficiently. This final approach is tested and implemented in the present paper.

A degree of uncertainly is associated with stable semantics. The standard algorithm for computing stable semantics is due to Modgil and Caminada [17]. As pointed out in their paper, the execution of the algorithm is terminated once an argument labeled *undec* is found. The uncertainly then stems from the fact that the run time of the algorithm depends primarily on the choice of the first argument for processing, which is done randomly. This feature occurs as much in computation with splitting as it does without splitting and has to be taken into consideration during the evaluation of the experimental data.

## 3. SPLITTING

Splitting consists in dividing an argumentation framework into two parts so as to perform semantical calculations separately on each part. After computing a partial extension/labeling on one part, a series of modifications are required before further computation can be carried out on the other part of the framework. Both outputs are then combined into one extension/labeling, thus producing the required result. Both the division and the required modifications depend on the type of splitting used.

### 3.1. Unidirectional Splitting

A unidirectional splitting procedure given in [12] requires that after dividing an AF all attacks lying between the two parts have their source in one part and their destination in the other. Formally:

International Journal of Artificial Intelligence and Applications (IJAIA), Vol.9, No.4, July 2018

**Definition 3**(Unidirectional Splitting [12])

Let $F_1 = (A_1, R_1)$ and $F_2 = (A_2, R_2)$ be argumentation frameworks such that $A_1 \cap A_2 = \emptyset$. Let $R_3 \subseteq A_1 \times A_2$. The triple $(F_1, F_2, R_3)$ is called a *unidirectional splitting* of the argumentation framework $F_0 = (A_1 \cup A_2, R_1 \cup R_2 \cup R_3)$.

A unidirectional splitting can be achieved e.g. by identifying the SCCs of the underlying directed graph. This approach appears intuitive considering the research conducted by Baroni et al [24] which indicates that all admissibility-based semantics proposed in [1], including the stable semantics, are SCC-recursive.

Tests conducted on argumentation frameworks using this kind of splitting are presented in [12]. The results demonstrate an improvement in run time by 54% on average for stable semantics. The unidirectional splitting as mentioned before works only for frameworks consisting of at least two SCCs. For frameworks consisting of a single SCC, unidirectional splitting has no effect.

In fact, random tests performed on argumentation frameworks of various complexity using the developed Java standalone application suggest that for 10 arguments and starting at roughly 30 attacks (i.e. 30% of all possible attacks for a framework of that size) the resulting framework will almost always constitute a single SCC. This ratio tends to decrease with the size of the framework. For 50 arguments around 300 attacks (12% of all possible attacks) were needed, and for 500 arguments around 3000 attacks (1.2% of all possible attacks).

### 3.2. Bidirectional Splitting

Unlike unidirectional splitting, bidirectional splitting is applicable to frameworks with an arbitrary structure.

The splitting procedure given in [14] consists in the following steps:
(1) find a cut $(F_1, F_2, R_3)$ of the framework $F_0$
(2) modify $F_1$ in accordance with Algorithm 1 (here we present our pseudocode based on the theory given in [14])
(3) compute the stable labelings of $F_1$
(4) for each extension obtained in step 3, modify $F_2$ and compute the stable labelings (see Algorithm 2, our pseudocode, for details see [14])
(5) combine the labelings of $F_1$ and $F_2$

---
**Algorithm 1: Bidirectional Splitting - part 1**
input : sub-framework $F_1$, $R_3$
output: modified framework $F_1^M$

1 **PROCEDURE** $assumeAttacked(F_1, R_3)$
2 **begin**
3     foreach $attack\ (b, a) \in R_3$ do
4         if $a \in A_1$ and $.a \notin A_1$ then
5             add $.a$ to $A_1$
6             add $(a, .a)$ to $R_1$
7             add $(.a, a)$ to $R_1$
8     return $F_1^M$

---

Prior to applying the splitting algorithm, the framework has to be divided into two sub-frameworks (Step 1 above). As proposed in [14], a minimum cut is desirable since the splitting algorithm will introduce additional arguments and attacks to both sub-frameworks $F_1$ and $F_2$. The introduced elements may lead to the creation of additional stable labelings (that will be discarded later), thus extending the time of the computation. The augmentation involves the following:

(1) In relation to $F_1$ (before a cut): for every attack $(b, a)$ which lies on the cut line ($R_3$) and whose target is in $F_1$, one additional argument and two attacks have to be inserted into $F_1$ (see Algorithm 1). This is to account for the possibility that the given argument may be attacked by $F_2$. If $a$ is in the extension of $F_1$ but is attacked by $b$ we know that $b$ cannot be in the extension of $F_2$ as it has to be labeled *out*.

(2) In relation to $F_2$ and an extension $E$ of $F_1$ (after a cut): for every attack $(b, a)$ which lies on the cut line ($R_3$) and whose target is in $F_1$ but not in $E$, an additional argument and a self-loop for this argument are added to $F_2$. This creates the possibility for argument $b$ to be in an extension of $F_2$. If on the other hand $a$ is in $E$, the source of the attack obtains a self-loop in $F_2$.

---
**Algorithm 2:** Bidirectional Splitting - part 2

**input** : sub-framework $F_2$, extension $E$ of $F_1^M$

**output:** modified framework $F_2^M$

1 **PROCEDURE** $removeAttacked(F_2, E)$
2 **begin**
3    **foreach** $attack\ (b, a) \in R_3$ **do**
4       **if** $a \in A_1\ and\ a \notin E$ **then**
5          **if** $: a \notin A_2$ **then**
6             $add : a$ to $A_2$
7             $add\ (:a, :a)$ to $R_2$
8          $add\ (b, :a)$ to $R_2$
9       **if** $b \in E$ **then**
10          $remove : a$ from $A_2$
11       **if** $a \in E$ **then**
12          $add\ (b, b)$ to $R_2$
13    **return** $F_2^M$

---

In order to minimize this overhead, it is advised in [14] to keep the number of attacks from $F_2$ to $F_1$, referred to as parameter $k$, minimal. As stated by the authors, $k$ arguments and $2k$ attacks would have to be added to $F_1$, and at most a further $k$ arguments and $2k$ attacks to $F_2$.

For minimum cut, we implement the Hao-Orlin (HO) algorithm ([25]) which is a flow-based procedure known for its good performance. As given in Chekuri et al ([26]), the time complexity of this algorithm is $O(|V| \times |E| \times \log(|V|^2/|E|))$, in which $V$ and $E$ stand for vertices and edges respectively.

We also develop a size-based approach, which we call *balanced cut*. It consists in dividing the framework into two sub-frameworks of equal size. The algorithm begins with initializing $A_1$ as an empty set and by adding an arbitrary argument to it. Gradually, the argument's neighbors are added one by one to $A_1$ as long as the size of $A_1$ is less than half the size of the framework. Since only half the arguments (and only the incoming attacks to those arguments) are processed, this algorithm has a linear time complexity. Algorithm 3 presents the pseudocode.

**Example 2**: Some possible bidirectional splittings for HO and BC are shown in Figure 2. For $HO_1$, $F_1 = \{3,4\}$, $F_2 = \{0,1,2\}$, and $k = 0$. For $HO_2$, $F_1 = \{4\}$, $F_2 = \{0,1,2,3\}$, and $k = 0$. For $BC_1$, it is either $F_1 = \{0,3,4\}$, $F_2 = \{1,2\}$ or $F_1 = \{1,2\}$, $F_2 = \{0,3,4\}$. In the latter two cases, parameter $k = 1$.

---

**Algorithm 3:** Balanced Cut Algorithm
**input** : argumentation framework $F$
**output**: $A_1$

1 **PROCEDURE** $balancedCut(F)$
2 **begin**
3      add an arbitrary argument $a$ to $A_1$
4      mark $a$ as "visited" and add it to $nextRound$
5      **while** $|A_1| < \frac{|A|}{2}$ **do**
6          $cut(nextRound)$
7      **return** $A_1$

8 **PROCEDURE** $cut(thisRound)$
9 **begin**
10      **foreach** $a \in thisRound$ **do**
11          remove $a$ from $nextRound$
12          **foreach** $at \in \{(i,a)\}$ **do**
13              **if** $i$ not visited and $|A_1| < \frac{|A|}{2}$ **then**
14                  mark $i$ as "visited", add $i$ to $A_1$ and to $nextRound$
15              **else**
16                  break
17      **if** $nextRound$ is empty **then**
18          add next unvisited argument $b$ to $nextRound$
19          add $b$ to $A_1$ and mark it as "visited"

---

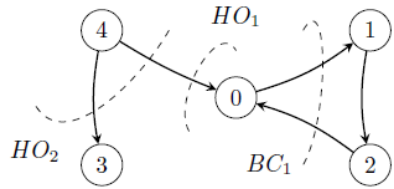

Figure 2: Example bidirectional splittings of a framework

## 4. EXPERIMENTAL SETUP

The experiment involves a sampling of 160 randomly generated frameworks resulting from extracting 20 instances from each of the following argument/attack combinations: 10/30, 10/50, 10/70, 10/80, 15/40, 20/40, 20/50, and 20/60. The random frameworks are generated as follows: after specifying the number of arguments and attacks, our algorithm utilizes Java's *Math.random()* function to perform the selection of each attack's source and target argument. This function returns a random real number from the interval [0,1]. Furthermore, we test also 5 complete frameworks containing $x \in \{6,7,8,9,10\}$ arguments and respectively $x^2$ attacks.

The focus is on frameworks where the number of attacks lies between $2x$ and $x^2$, with $x$ being the number of arguments. This choice is dictated by the fact that unidirectional splitting works well for frameworks of at least two SCCs, which are rather sparse frameworks where the number of attacks is around or less than double the number of arguments, as previously shown in [12]. Thus, it is reasonable to concentrate on frameworks with a more complex structure, which in general leads to frameworks consisting of just a single SCC.

All test algorithms are implemented using the Java programming language and are constituent parts of our standalone application, on wich the exeperimental evaluation is performed. The implementations include the Hao-Orlin minimum cut algorithm (HO, [25]), the bidirectional splitting algorithm ([14]), the stable semantics algorithm ([17]), and the balanced cut algorithm (BC) that we propose.

The empirical evaluation is performed on two computers, both running Windows 10 64-bit. Computer A has an Intel(R) Pentium(R) CPU P6200 2.13 GHz processor with two physical and two logical cores and 4 GB RAM. Computer B has a dual core Intel(R) Core(TM) i5-5200U 2.20 GHz processor with four logical cores and has 8 GB RAM. In the following we refer to the two computers per their RAM size, i.e. 4 GB or 8 GB.

The run-times are given in milliseconds. 0 ms indicates that the execution time is below 1 ms. Percentage values are rounded to the nearest integer according to the common rounding convention: round up in case of .5 or above and round down if otherwise. A value of 100% indicates that the gain in time is above 99.5%. A minus sign in front of a value for gain in time indicates an execution that takes longer with the application of splitting than the respective execution without splitting.

The trial phase of the experiment has determined that for 20 arguments and 60 attacks the computation of stable semantics without splitting can reach well above 60 minutes. For reference, the computation of stable semantics for the 10/100 AF on 4 GB RAM using splitting with HO was terminated after 3 hrs without completing the calculation. Hence, a time limit for the measurement of execution time of 30 minutes or 1,800,000 ms is imposed. All runs that exceed this limit are terminated and marked correspondingly in the results with a ">" sign preceding the value. It should be stressed at this point that we are primarily interested in comparing the run times for bidirectional splitting with the run times without splitting rather than the execution times *per se*.

The experiments consist in executing the algorithms six times on each framework - once for computation without splitting, once for computation with HO minimum cut, and once for computation with BC - and each time either with 4 GB or 8 GB RAM. Given that 165 frameworks were tested, a total of 990 tests were conducted.

## 5. RESULTS AND DISCUSSION

The first four sets of tests are conducted on frameworks with 10 arguments and 30, 50, 70, 80 attacks respectively. The purpose is to determine the behavior of the algorithms given a fixed number of arguments and an increasing number of attacks.

Test results for 10/30 and 10/50 frameworks are plotted in Figure 3. In the case of 10/30 AFs, splitting with both HO minimum cut and BC is faster than the execution without splitting, with BC introducing a significant speedup. The average run time without splitting is 768 ms and 255 ms for 4 GB and 8 GB, respectively. In the case of HO it is 235 ms and 155 ms with an average speedup of 50% and 41% for 4 GB and 8 GB, respectively. In the case of BC the average execution time takes 24 ms and 4 ms and leads to an improvement by 90% and 88% for 4 GB and 8 GB, respectively.

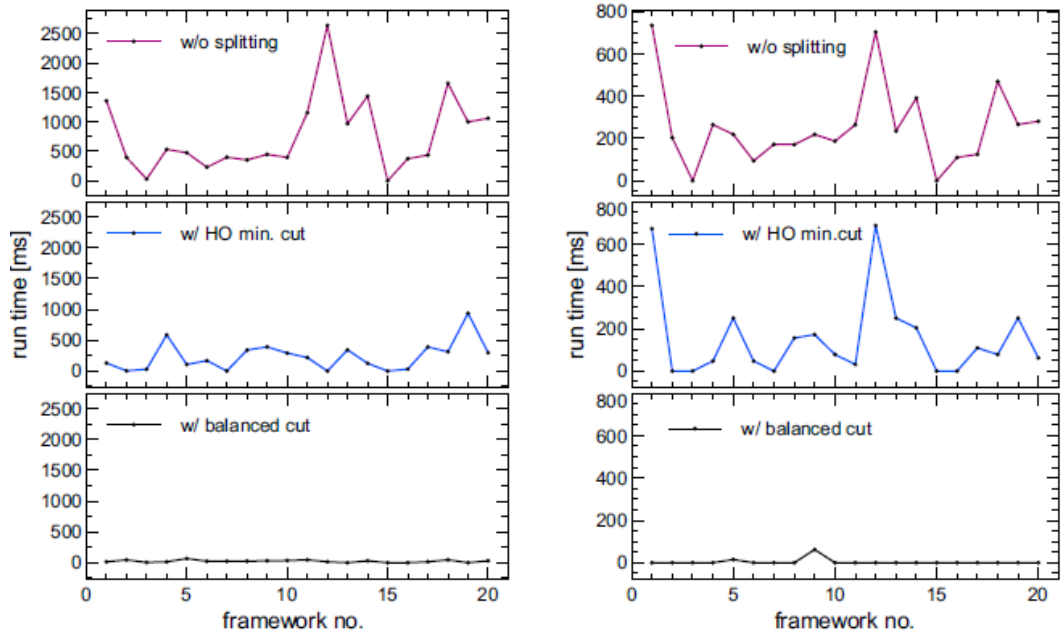

(a) framework type: 10/30, 4 GB          (b) framework type: 10/30, 8 GB

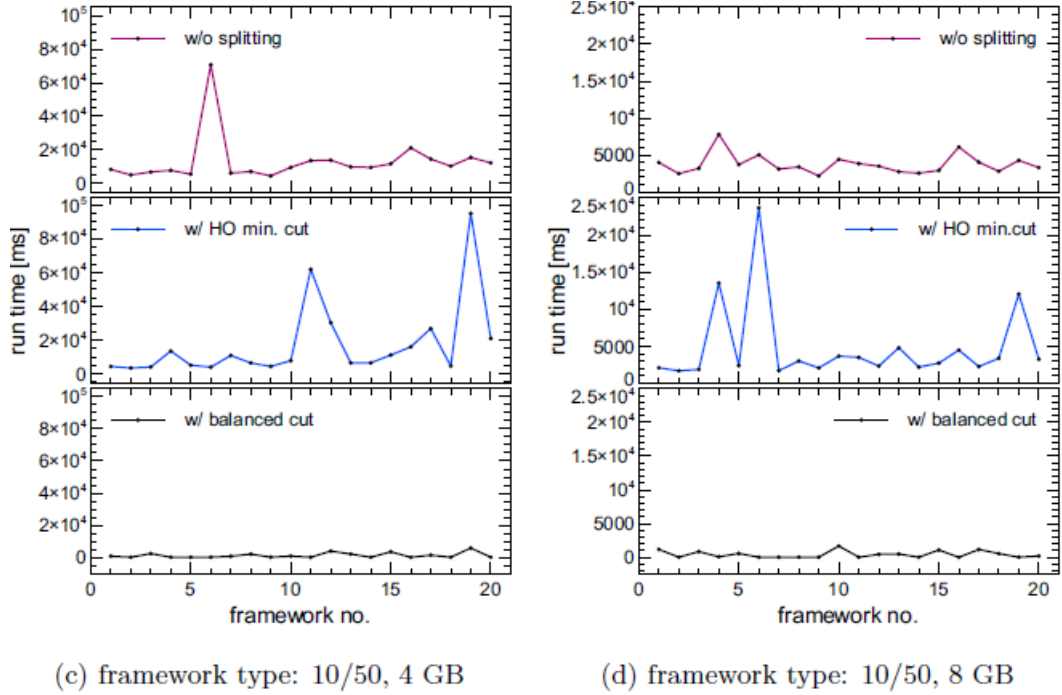

Figure 3: Execution times for frameworks with 10 arguments and 30 or 50 attacks, respectively.

For frameworks with 10 arguments and 50 attacks we observe a decrease in performance for splitting with HO. In fact, it is on average slower than execution without splitting, in both 4 GB and 8 GB cases. The performance of splitting with BC stays unchanged, with an average speedup of 89% and 88% for 4 GB and 8 GB, respectively.

Figure 4 shows the run time for frameworks with 10 arguments plus 70 and 80 attacks, respectively. Here, the performance with HO deteriorates further as compared to both executions without splitting and with BC splitting.

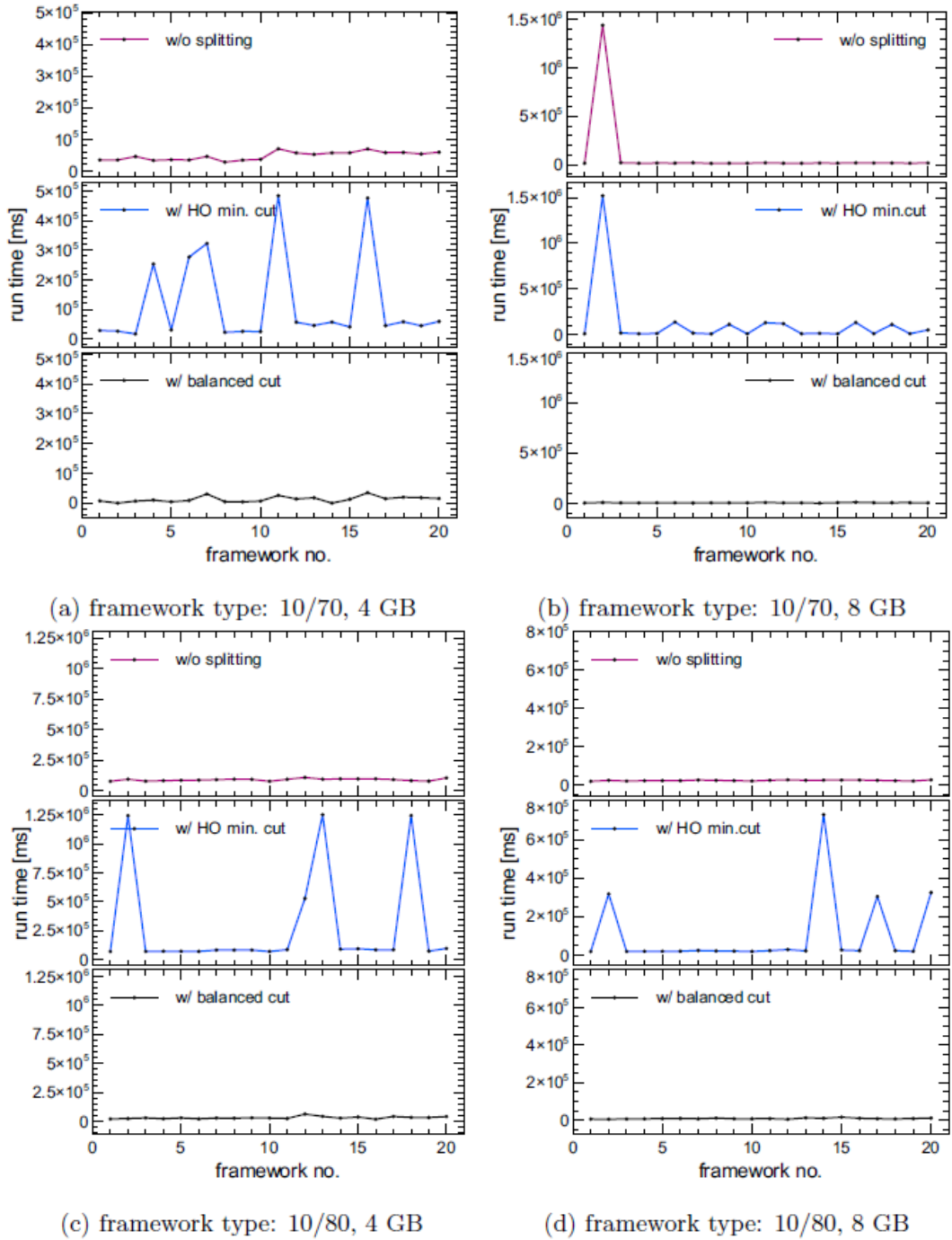

Figure 4: Execution times for frameworks with 10 arguments and 70 or 80 attacks, respectively.

For the 10/70 frameworks, the average run time without splitting is 47.5 and 87.5 sec for 4 GB and 8 GB, respectively. Although in general a faster performance is recorded for 8 GB RAM than for 4 GB, the discrepancy here showcases the uncertainty of computing the stable semantics which depends on how fast the first element labeled *undec* can be found. Once such an element is found, the computation is terminated as an *undec* element indicates that no stable labeling exists.

In the case of BC, the average gain in time as compared to no splitting is 74-75% for the 10/70 AFs and 66-67% for the 10/80 AFs.

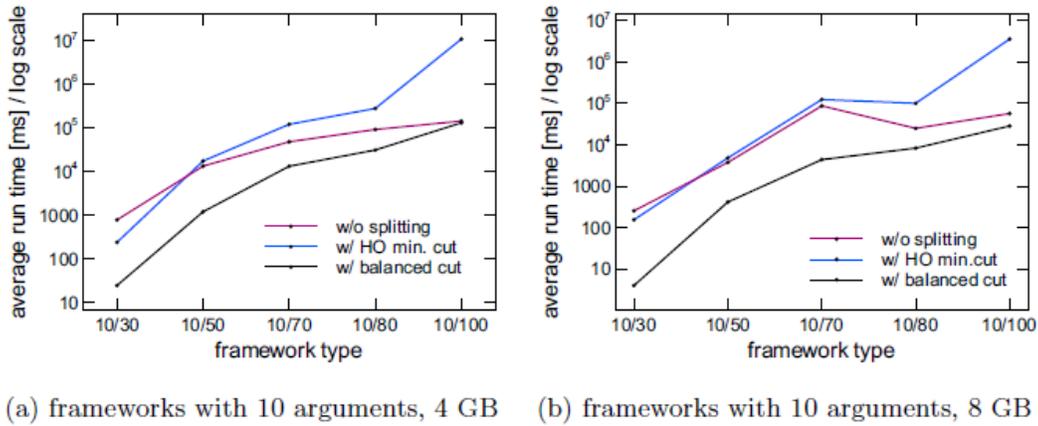

(a) frameworks with 10 arguments, 4 GB  (b) frameworks with 10 arguments, 8 GB

Figure 5: Summary of average execution times across frameworks with 10 arguments and 30; 50; 70; 80; or 100 attacks.

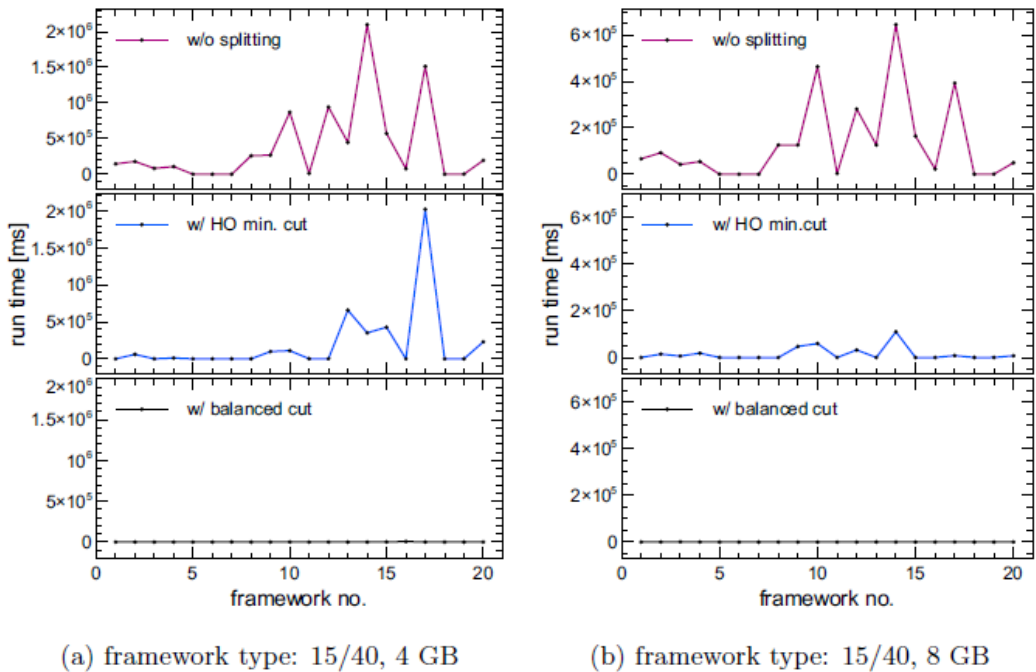

(a) framework type: 15/40, 4 GB  (b) framework type: 15/40, 8 GB

Figure 6: Execution times for frameworks with 15 arguments and 40 attacks

Average run time values for frameworks with 10 arguments are summarized in Figure 5. We include in it the results for the 10/100 framework, which has the maximum number of attacks for 10 arguments. Splitting with HO minimum cut performs on average better only for the 10 arguments and 30 attacks combination. For the remaining framework types, its execution time lies well above the other two methods. We observe a good improvement for splitting with BC over the entire range of framework size for frameworks with 10 arguments.

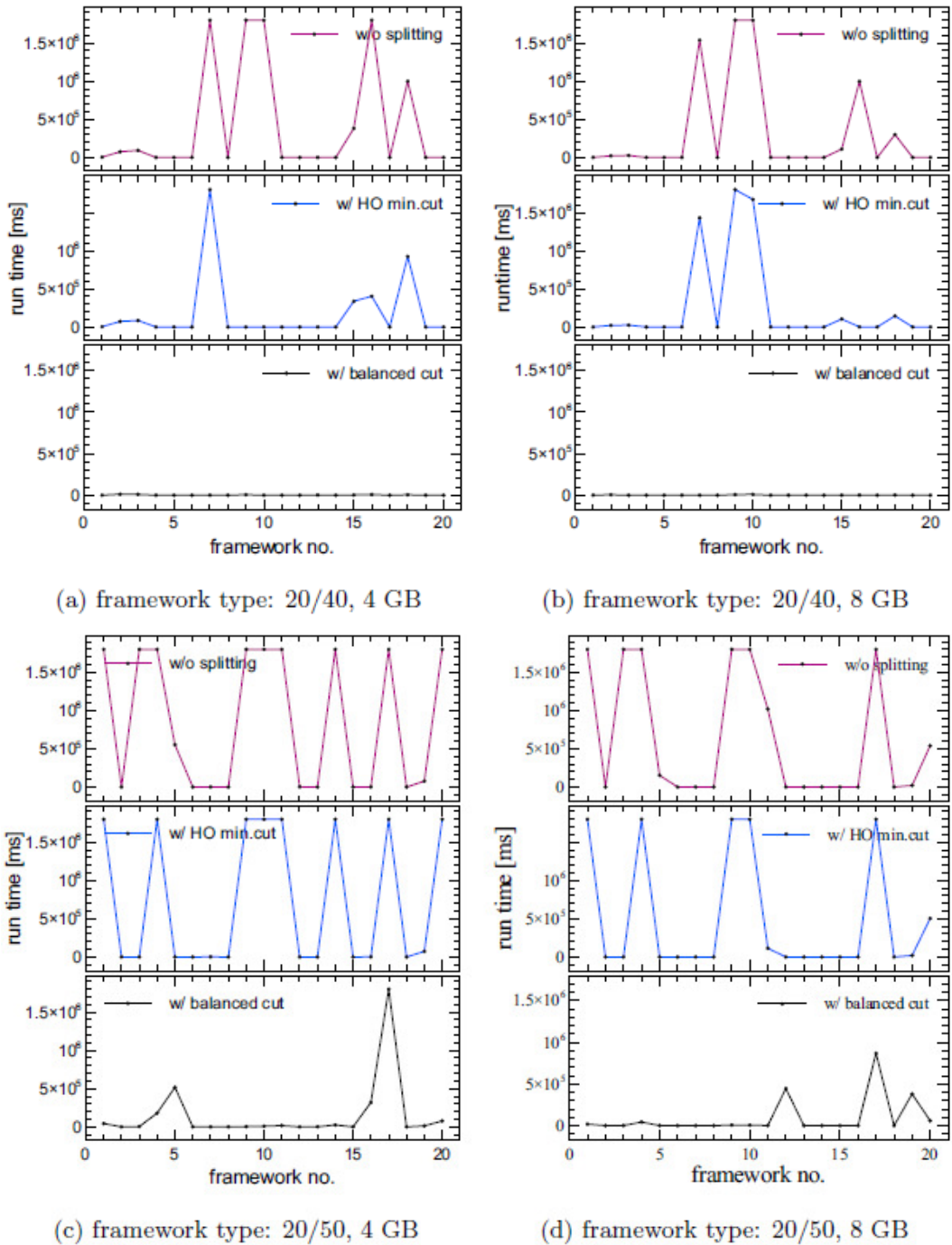

Figure 7: Execution times for frameworks with 20 arguments and 30 or 50 attacks, respectively.

The run time values for the combination of 15 arguments and 40 attacks are shown in Figure 6. In cases without splitting, the average execution time reaches 6.5 and 2.2 min, respectively for 4 GB and 8 GB. For splitting with HO, on average a performance of 3.3 and 0.2 min is recorded for 4 GB and 8 GB, respectively. The average execution time for splitting with BC is a mere 15 and 0.2

23

sec., respectively for 4 and 8 GB. These results are consistent with the data we obtained for frameworks with 10 arguments.

Figure 7 shows the run times for frameworks of 20 arguments plus 40 and 50 attacks, respectively. The speed-up achieved by splitting with BC is best also for these two types of AFs. The performance of splitting with HO is only slightly better than that without splitting. The same holds for frameworks with 20 arguments plus 60 attacks as shown in Figure 8.

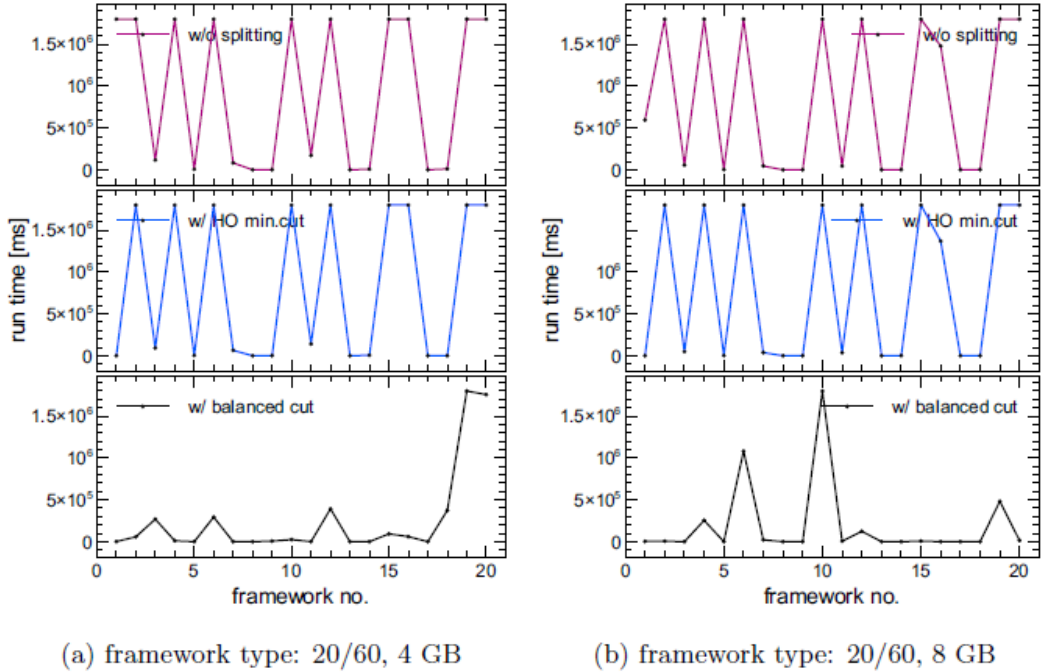

(a) framework type: 20/60, 4 GB    (b) framework type: 20/60, 8 GB

Figure 8: Execution times for frameworks with 20 arguments and 60 attacks.

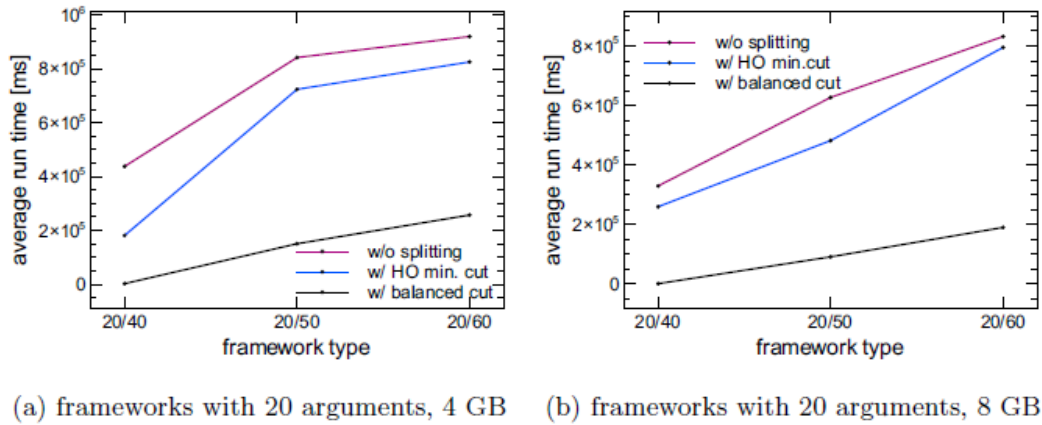

(a) frameworks with 20 arguments, 4 GB    (b) frameworks with 20 arguments, 8 GB

Figure 9: Summary of average execution times across frameworks with 20 arguments and 40; 50; or 60 attacks.

An overview of results for AFs with 20 arguments is plotted in Figure 9. As in the case of frameworks with 10 arguments, notable is an improvement in performance of splitting with BC the larger the number of attacks gets.

24

Based on the conducted tests, we make the observation that the computation time for all three methods in general increases with both the number of arguments and the number of attacks. However we are able to show that, for a given number of arguments, the performance of splitting with BC gets better with an increasing number of attacks. It is to the contrary for splitting with HO, i.e., the performance decreases as the number of attacks becomes higher.

This is supported further by the results we obtain for AFs with $x \in \{6,7,8,9,10\}$ arguments and respectively a maximal number of $x^2$ attacks (Figure 10). In fact, for these frameworks, HO never performs better than execution without splitting nor with splitting with BC. On the other hand, splitting with BC is approximately twice as fast as when no splitting is applied. Furthermore, the data demonstrate a tendency for BC to perform better as the framework becomes larger. Consider runs on 8 GB, for example. BC registers an improvement of 33% for the framework 7/49, 40% for 8/64, 45% for 9/81 and the time gain is 50% for 10/100.

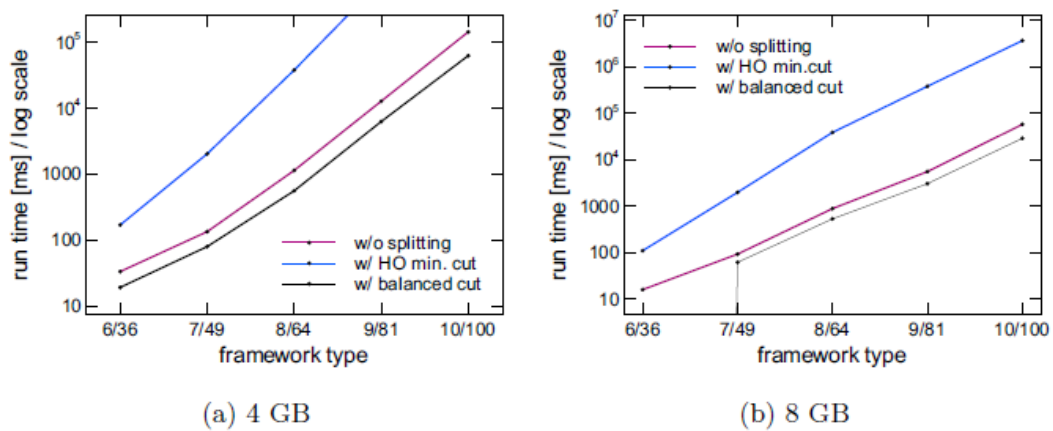

Figure 10: Execution times for frameworks with x arguments and $x^2$ attacks, where $x \in \{6, 7, 8, 9, 10\}$. The run with HO minimum cut is terminated after 3 hr for 4 GB, after 1 hr for 8 GB without conclusion.

Based on our experimental results we expect that the run time values and gain values obtained for smaller frameworks will scale. This is supported especially by the performance across 10 (or 20) arguments and varying number of attacks as shown in Figure 5 and Figure 6.

A summary of the average execution times and time gains for all types of randomly generated frameworks that we have tested is given in Table 1 for 4 GB and in Table 2 for 8 GB. In both, a discrepancy between the run times and the gain in time for both HO and BC is noticeable. Take for instance the 15/40 frameworks in the 4 GB case. Here, the average run time without splitting is 388,638 ms. When splitting with HO is used, the average run time is reduced to 198,774 ms, which in terms of value corresponds to an improvement of 49% as compared to the average run time without splitting. When the average gain is calculated by averaging over time gain per each of the 20 frameworks, it amounts to 52%, which is relatively close to 49%. The situation is quite different for BC. Here, the average run time is 15,378 ms, which corresponds to an average improvement of 96% in terms of value. However, when the average gain is calculated by averaging over all 20 time gains, the result is a deceleration of 2,148%. The discrepancy is significant and it results from the fact that in 2 out of 20 frameworks the computation of the stable labeling with BC took longer than without splitting, 504% and 44,080%, respectively. It should be noted though that this behavior is highly irregular as splitting with BC is 100% faster for as many as 15 out of 20 tested frameworks. This is the effect of the inherent uncertainty in computing stable semantics. This is also why we primarily focus on the average execution times instead of the percentage gain.

Table 1: Summary of average results for random frameworks: run time and time gain. (4 GB) Noticeable is a discrepancy between the average run times and gains.

| AF type | w/o spl. [ms] | HO [ms] | gain [%] | BC [ms] | gain [%] |
|---|---|---|---|---|---|
| 10/30 | 768 | 235 | 50 | 24 | 90 |
| 10/50 | 13178 | 17220 | -45 | 1181 | 89 |
| 10/70 | 47539 | 119563 | -144 | 13064 | 74 |
| 10/80 | 91023 | 276641 | -200 | 30775 | 66 |
| 15/40 | 388638 | 198774 | 52 | 15378 | -2148 |
| 20/40 | > 437688 | > 181349 | -180 | 2528 | -1726 |
| 20/50 | > 841798 | > 723820 | -205 | > 150774 | -8159 |
| 20/60 | > 919803 | > 825392 | -100 | > 257202 | -233 |

Table 2: Summary of average results for random frameworks: run time and time gain. (8 GB) Noticeable is a discrepancy between the average run times and gains.

| AF type | w/o spl. [ms] | HO [ms] | gain [%] | BC [ms] | gain [%] |
|---|---|---|---|---|---|
| 10/30 | 255 | 155 | 41 | 4 | 88 |
| 10/50 | 3773 | 4847 | -18 | 417 | 88 |
| 10/70 | 87500 | 124724 | -199 | 4399 | 75 |
| 10/80 | 25109 | 101079 | -288 | 8344 | 67 |
| 15/40 | 132875 | 15245 | 81 | 205 | 90 |
| 20/40 | > 329638 | > 259880 | 15 | 1499 | -3942 |
| 20/50 | > 626801 | > 481914 | -76 | 91533 | -6311 |
| 20/60 | > 831480 | > 794871 | 26 | > 190299 | 74 |

Data obtained in the experiment suggest that bidirectional splitting leads to a significant improvement in computing stable semantics in argumentation frameworks if it is applied to frameworks with a single SCC.

The computational overhead resulting from the bidirectional splitting method proposed in [14] can be overcome by using BC instead of HO. Furthermore, BC has yet another advantage over a minimum cut such as HO. We have seen that the execution times using BC are more evenly spread out. The plots for different types of frameworks have shown that splitting with BC rarely results in very high values which seem disproportionate to the other outcomes. This also means that our method is able to predict, to a certain extent, the run time. This is not possible when HO minimum cut is used as the values obtained this way may range from very small to very large (e.g. in the case of the 20/60 frameworks, 9 of them exceeds the 30 min limit with 4 GB and 8 with 8 GB whereas the rest lies well below that time length).

A further research into the viability of splitting would be to test the execution times on real-life argumentation systems.

Our experimental data confirm the statement by Dunne and Wooldridge that processing a stable semantics is computationally challenging and resides in the class of NP-complete problems. Hence developing methods that speed up the process even further is imperative. We expect a possible improvement in the possibility of tapping into the power of the superposition principle of quantum computing. To this end, more research into how non-regular directed graphs can be represented by Hermitian adjacency matrices is needed for the evolution of the quantum system to be unitary. A work on symmetric argumentation frameworks - the type that corresponds to a Hermitian matrix with real entries - conducted by Coste-Marquis et al [27] shows that AFs with symmetric attack relation always possess at least one stable extension.

## 6. CONCLUSIONS

Our experimental results demonstrate that bidirectional splitting has a good potential to speed up the computation of stable semantics in argumentation frameworks. But if we want the speedup to be reliable at all, the balanced cut method should be used instead of a minimum cut proposed in [14], as the results suggest that the $k$ parameter does not play a role in the speedup. Then if bidirectional splitting with a minimum cut (be it HO style or others) is used, the improvement in performance is recorded for rather sparse frameworks, which consist of at least two SCCs. It simply means that the performance of bidirectional splitting with a minimum cut works as well as that of unidirectional splitting (achieving an average speedup of 54%) in speedup for sparse frameworks.

The data suggest using unidirectional splitting for rather sparse frameworks (of up to approximately 2.5 attacks per argument) is desirable as this type of frameworks rarely constitute a single strongly connected component. For denser AFs (starting at roughly 2.5 attacks per argument and onward), which in most cases consist of just one single SCC, we propose the use of bidirectional splitting with our BC method. In disputation terms we can treat those denser AFs as being contentious.

The computation of stable semantics may reach high run times of lengths not easily predictable due to the lack of regularity. Splitting in such a way that the underlying graph is divided into two sub-graphs of equal magnitude leads to execution times that are more evenly spread out and spikes in values are seldom observed. Splitting with BC is therefore a more favorable method with regard to the prediction of execution time.

A word of explanation is due as to why the application of the HO minimum cut to bidirectional splitting did not demonstrate a significant speedup in computing stable semantics. It is partly due to the requirement of keeping the parameter $k$ minimal. The requirement in fact induces a bias towards unidirectionality. Indeed, if $k=0$ then the underlying graph consists of more than one SCC, and the semantics is thus better computable with unidirectional splitting applied.

**ACKNOWLEDGEMENTS**

Part of the work for the present paper was conducted at Leipzig University, Germany and most of the work at Nanjing University, China.